\documentclass[journal]{IEEEtran}

\usepackage{caption}
\usepackage{booktabs}  
\usepackage{siunitx}
\usepackage{amsmath, amssymb, amsfonts, amsthm}
\usepackage{algorithm}
\usepackage{algorithmic}
\usepackage{cite}
\usepackage{amsthm}

\usepackage{lipsum}
\usepackage{color}
\usepackage{multirow}
\usepackage{graphicx}
\usepackage{float}
\usepackage{hyperref}

\usepackage{threeparttable}

\ifCLASSOPTIONcompsoc \usepackage[caption=false,font=normalsize,labelfont=sf,textfont=sf]{subfig}
\else
\usepackage[caption=false,font=footnotesize]{subfig}
\fi
\hypersetup{hidelinks,
	colorlinks=true,
	allcolors=black,
	pdfstartview=Fit,
	breaklinks=true}

\begin{document}

\title{TransMPC: Transformer-based Explicit MPC with Variable Prediction Horizon}

\author{Sichao~Wu,
        Jiang~Wu,
        Xingyu~Cao,
        Fawang~Zhang,
        Guangyuan~Yu,
        Junjie~Zhao,
        Yue~Qu,
        Fei~Ma,
       and Jingliang~Duan*
\thanks{All correspondences should be sent to J.Duan with email: duanjl@ustb.edu.cn.}
\thanks{S. Wu, J. Wu, X. Cao, G. Yu, J. Zhao, F. Ma and J. Duan are with the School of Mechanical Engineering, University of Science and Technology Beijing, China, 100083. {\tt\small Email: \url{wsc@xs.ustb.edu.cn}, \url{wujiang0826@xs.ustb.edu.cn}, \url{caoxingyu2022@xs.ustb.edu.cn}, \url{M202410461@xs.ustb.edu.cn}, \url{zhaojunjie@xs.ustb.edu.cn}, \url{yeke@ustb.edu.cn}, \url{duanjl@ustb.edu.cn}}.}
\thanks{F. Zhang is with the School of Mechanical Engineering, Beijing Institute of Technology, Beijing, China, 100081. {\tt Email:\url{fawangzhang@bit.edu.cn}}.}
\thanks{Y. Qu is with the School of Mechanical Engineering, Nanjing University of Industry Technology, Nanjing 210023, China. {\tt Email:\url{2024101520@niit.edu.cn}}.}
}
\maketitle

\begin{abstract}
Traditional online Model Predictive Control (MPC) methods often suffer from excessive computational complexity, limiting their practical deployment. Explicit MPC mitigates online computational load by pre-computing control policies offline; however, existing explicit MPC methods typically rely on simplified system dynamics and cost functions, restricting their accuracy for complex systems. This paper proposes TransMPC, a novel Transformer-based explicit MPC algorithm capable of generating highly accurate control sequences in real-time for complex dynamic systems. Specifically, we formulate the MPC policy as an encoder-only Transformer leveraging bidirectional self-attention, enabling simultaneous inference of entire control sequences in a single forward pass. This design inherently accommodates variable prediction horizons while ensuring low inference latency. Furthermore, we introduce a direct policy optimization framework that alternates between sampling and learning phases. Unlike imitation-based approaches dependent on precomputed optimal trajectories, TransMPC directly optimizes the true finite-horizon cost via automatic differentiation. Random horizon sampling combined with a replay buffer provides independent and identically distributed (i.i.d.) training samples, ensuring robust generalization across varying states and horizon lengths. Extensive simulations and real-world vehicle control experiments validate the effectiveness of TransMPC in terms of solution accuracy, adaptability to varying horizons, and computational efficiency.
\end{abstract}


% The paper headers
\markboth{Journal of \LaTeX\ Class Files,~Vol.~14, No.~8, August~2015}%
{Shell \MakeLowercase{\textit{\emph{et al.}}}: Bare Demo of IEEEtran.cls for IEEE Journals}
% \IEEEpeerreviewmaketitle

\section{Introduction}

\IEEEPARstart{M}{odel} predictive control (MPC), as a real-time optimization control method, has been widely applied in fields such as industrial automation, autonomous driving, and drone control \cite{qin2003survey,vazquez2014model,li2014fast}. MPC demonstrates high flexibility and superiority in handling constraints, nonlinear systems, and multi-objective optimization problems. However, a major challenge with existing MPC algorithms lies in their relatively low computational efficiency \cite{lee2011model}. 

To address this issue, the move-blocking technique was introduced, which assumes that the control input remains constant within a fixed portion of the prediction horizon, thereby reducing the number of variables to be optimized and improving computational efficiency. However, this technique struggles to achieve its full potential while maintaining control performance, system stability, and constraint satisfaction \cite{cagienard2007move}. Wang and Boyd  \cite{wang2009fast} proposed an early termination interior-point method by limiting the maximum number of iterations per time step, further reducing computation time. Odonoghue et al. \cite{o2013splitting} utilized operator splitting methods to solve MPC problems. By employing the alternating direction method of multipliers, the problem is simplified. This approach enhances computational efficiency at the expense of some computational precision .

Despite these advances in online optimization, computational limitations remain a significant barrier for real-time implementation in complex systems. To further improve computational efficiency, explicit MPC emerged, which precomputes and stores the optimal control policies in advance, allowing for direct retrieval during online computations, thereby significantly reducing the computational burden. Bemporad et al. \cite{bemporad2002explicit} proposed the explicit MPC approach by dividing the state space into multiple regions and calculating a feedback control law for each region. During online operation, the appropriate control law is selected, but this method is only applicable to small-scale systems, as storage requirements grow exponentially with increasing state dimensions \cite{kouvaritakis2002needs}. To address this issue, various approximate MPC algorithms have been proposed. For example, Geyer et al. \cite{geyer2008optimal} reduced the number of polyhedral state regions by merging regions with identical control laws. Jones et al. \cite{jones2010polytopic} utilized dual descriptions and centroid functions to estimate optimal policies, significantly reducing the number of partitioned regions. Wen et al. \cite{wen2009analytical} employed piecewise continuous grid functions to represent the explicit MPC solution, reducing storage demands and enhancing online computational efficiency. Additionally, Borrelli et al. \cite{borrelli2010computation} proposed a hybrid explicit MPC algorithm that combines both online and offline execution. However, the abovementioned explicit MPC methods still struggle with high model dependency and increasing computational demands for complex systems. 

Recent progress in deep learning has enabled neural networks (NNs) to tackle complex temporal data and nonlinear systems effectively, facilitating their integration into MPC frameworks. Moreover, some MPC implementations use parameterized functions to approximate the controller, with parameters optimized via supervised or reinforcement learning to minimize the MPC cost function over a fixed prediction horizon \cite{aakesson2005neural,aakesson2006neural,cheng2015neural}. To further address the limitations of explicit MPC in real-time performance and computational efficiency for complex dynamic system control, Liu et al.~\cite{liu2022recurrent} proposed the Recurrent Model Predictive Control (RMPC) method. This method employs Recurrent Neural Networks (RNN) to approximate optimal control policies, significantly reducing online computational burden through offline policy learning and achieving dynamic adjustment of prediction horizons. However, it can only output single-step actions, making it impossible to reasonably evaluate our effective actions. Finite Horizon Approximate Dynamic Programming (FHADP \cite{li2023reinforcement}) addresses the issue of single-step action output by generating action sequences over the entire prediction horizon. However, it loses the flexibility to dynamically adjust the prediction horizon, thereby failing to adapt to the changing environment.

To address the above limitations, we present TransMPC, an explicit MPC framework that couples a Transformer encoder with a policy‐gradient training scheme.  TransMPC natively handles variable prediction horizons and produces high-accuracy control sequences in real time. The principal contributions of this study is summarized as:
\begin{enumerate}
    \item {We formulate the MPC policy as an encoder-only Transformer that maps the current state and a reference trajectory of arbitrary length to an entire open-loop action sequence in one forward pass.  In contrast to existing explicit MPC methods constrained by fixed prediction horizons~\cite{li2023reinforcement}, the bidirectional attention mechanism of the policy inherently supports varying horizons and maintains low inference latency.}
    \item  We develop the TransMPC algorithm, which alternates between two phases: sampling and learning.  Unlike imitation-learning approaches that rely on mimicking precomputed optimal trajectories~\cite{celestini2024transformer,zinage2024transformermpc}, our method directly minimizes the finite-horizon MPC cost through automatic differentiation. A replay buffer combined with random horizon sampling ensures that the learned policy generalizes effectively across the entire state space and different prediction horizons.
    \item Simulation and real-world experiments  demonstrate that TransMPC attains an order-of-magnitude speed-up over classical on-line MPC solvers~\cite{Andreas2006Biegler,bonami2008algorithmic}, while achieving higher control accuracy than prior explicit MPC approaches based on MLP or RNN architectures~\cite{li2023reinforcement,liu2022recurrent}.
\end{enumerate}

The paper is organized as follows: Section \ref{sec: preliminary} formalizes the MPC problem; Section \ref{sec: method} analyzes the characteristics of different network structures in detail and introduces the core components and theoretical foundations of the TransMPC algorithm, demonstrating Transformer's advantages in handling variable-length prediction horizons; Section \ref{sec: simulation validation} comprehensively evaluates the performance of the TransMPC algorithm through trajectory-tracking simulation experiments, verifying the effectiveness of the TransMPC framework; finally, Section \ref{sec: conclusion} concludes the paper and suggests directions for future research.

\section{Preliminary}
\label{sec: preliminary}
\subsection{Finite Horizon Optimal Control Task}
\label{subsec: Finite Horizon Optimal Control Task}

For a general time-invariant discrete-time dynamic system, the N-step MPC problem without state constraints is defined:
\begin{equation}
\label{eq:optimal_control_function}
\begin{aligned}
&\underset{u_t,\dots,u_{t+N-1}}{\min} V(x_t, X^R, N) = \sum_{i=0}^{N-1} l(x_{t+i}, x_{t+i}^R, u_{t+i})\\
&\mathrm{s.t.} \quad x_{t+1}=f(x_{t},u_{t}), 
\end{aligned}
\end{equation}
where \( V(x_t, X^R, N) \) is the cost function from the initial state \( x_t \in \mathcal{X} \subset \mathbb{R}^{n_{\rm state}} \), \( t \) is the initial time, \( N \) is the horizon length, \( X^R = [x^R_{t},\cdots,x^R_{t+N-1}]^\top\subset \mathbb{R}^{N\times n_{\rm ref} } \) is the reference signal sequence, \( u_{t+i} \in \mathcal{U} \subset \mathbb{R}^{n_{\rm input}}\) is the control input of the \( i \)-th step, \( l \geq 0 \) is the running cost. The system dynamics function \( f : \mathbb{R}^{n_{\rm state}} \times \mathbb{R}^{n_{\rm input}} \to \mathbb{R}^{n_{\rm state}} \). We assume that \( f(x_{t}, u_{t}) \) is Lipschitz continuous and stabilizable on a compact set \( \mathcal{X} \).

The purpose of Model Predictive Control (MPC) is to determine the optimal control sequence that minimizes the objective function $V(x_t, X^R, N)$,which can be denoted as:

\begin{equation}
\label{eq:MPC_function}
\begin{aligned}
U^*(x_t,X^R)=
\mathop{\arg\min}_{u_t, \dots, u_{t+N-1}} V(x_t, X^R, N),
\end{aligned}
\end{equation}
where the superscript $^*$ denotes the optimum and $$U(x_t,X^R):=\left[ u_t, u_{t+1}, \dots, u_{t+N-1} \right]^\top.$$

\subsection{Policy Sequence Approximation}
One viable policy to alleviate the real-time computational burden of MPC is to learn an approximate mapping from the system state (and reference) to the optimal control sequence, and then deploy this mapping online. By offloading the bulk of optimization to an offline training phase, the online computational requirements can be significantly reduced. 
 
Let $\pi(x_{t},X^{R};\theta):\mathbb{R}^{n_{\mathrm{state}}}\times\mathbb{R}^{N\times n_{\mathrm{ref}}}\!\longrightarrow\!\mathbb{R}^{N \times n_{\mathrm{input}}}$
denote a policy network with parameters~$\theta$, mapping the current state~$x_{t}$ and the future reference signal~$X^{R}$ to an $N$‐step control sequence  
\(
U(x_{t},X^{R})
\).
Using this policy, the finite‐horizon optimal‐control problem~\eqref{eq:optimal_control_function} is recast as an explicit MPC training problem
\begin{equation}
\label{eq:problem2}
\begin{aligned}
&\underset{\theta}{\min}\  V(x_t, X^R, N;\theta) = \sum_{i=0}^{N-1} l(x_{t+i}, x_{t+i}^R, u_{t+i}^N(x_t, X^R))\\
&\mathrm{s.t.} \quad x_{t+1}=f(x_{t},u_{t})\\
&\qquad \pi(x_t,X^R;\theta)=U(x_t,X^R).
\end{aligned}
\end{equation}

\section{Methodology}
\label{sec: method}

This section introduces a novel explicit MPC framework, named TransMPC, that employs a Transformer encoder as its policy approximator.  TransMPC naturally supports variable prediction horizons and delivers real-time control sequences with minimal computational overhead.  The remainder of this section detail the core components of TransMPC, with particular emphasis on the policy function design and algorithm development.

\subsection{Encoder-based Policy Network}
\label{subsec: Encoder-based Policy Network}

Prior to approximating the optimal control sequence, the policy function
\(\pi(x_{t},X^{R};\theta)\) must possess sufficient expressive power to admit parameters $\theta^{\star}$ such that
\begin{equation}
\label{eq.optimal approximation}
    \pi(x_{t},X^{R};\theta^{\star}) \;\approx\; U^{\star}(x_{t},X^{R}),
\end{equation}
where $U^{\star}(x_{t},X^{R})$ is the optimal sequence defined in~\eqref{eq:MPC_function}.  
Neural networks, by virtue of their universal approximation capability, provide a suitable function class for this requirement.  
We therefore review the suitability of three representative network architectures for explicit MPC: multi-layer perceptrons, recurrent networks, and attention-based Transformers.

\subsubsection{Multi-Layer Perceptron (MLP) structures}
MLPs accept fixed-dimension inputs and emit fixed-dimension outputs; consequently, the prediction horizon~$N$ is hard-coded into the architecture. This inflexibility precludes their direct use in variable-horizon settings and, because MLPs lack an inductive bias for sequential structure, they are generally ineffective at capturing the long-range temporal dependencies that underpin optimal control trajectories.

\subsubsection{Recurrent structures}
Recurrent neural networks, including gated variants such as GRU, Mamba, and TTT \cite{8053243,gu2023mamba,sun2020test}, are designed for sequential data. Unidirectional RNNs summarise only the past; for MPC, this means the control sequence must be inferred solely from the final hidden state, incurring computational load. Bidirectional RNNs access both past and future context, but at the expense of doubled computational effort and increased latency. Moreover, the intrinsic step‐by‐step state update impedes parallelism and can limit throughput in real-time applications.

\subsubsection{Attention-based structures}
Transformers overcome these limitations by representing each token as a key–value pair and computing outputs via scaled dot-product attention. They therefore (i) accommodate variable-length inputs and outputs, (ii) expose substantial intra-sequence parallelism, and (iii) retain global context even for long horizons—features that are particularly attractive for explicit MPC.

\begin{figure}[!htbp]
    \centering
    \captionsetup{singlelinecheck = false, labelsep=period, font=small}
    \captionsetup[subfigure]{justification=centering}
    \includegraphics[width=0.3\textwidth]{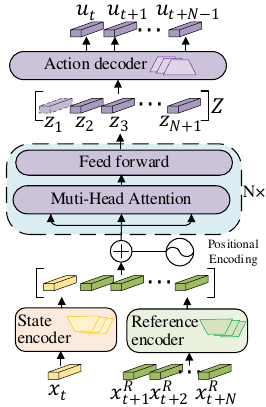}
    \caption{Transformer-based policy architecture for $N$-step action prediction. 
    The current state $x_t$ and future reference states $\{x_{t+1}^R,\dots,x_{t+N}^R\}$ are separately encoded, concatenated, and enriched with positional information. 
    Stacked multi-head attention and feed-forward layers transform the sequence into latent vectors $\{z_1,\dots,z_{N+1}\}$, which an action decoder maps to the control sequence $\{u_t,\dots,u_{t+N-1}\}$.}
    \label{fig:tf_policy}
\end{figure}

Exploiting the strengths of attention-based structures, we adopt an encoder-only Transformer  as the policy approximator (as shown in Fig.~\ref{fig:tf_policy}):  
\begin{equation}
\label{eq:policy}
\begin{aligned}
&\pi(x_{t},X^{R};\theta)\;=\;
D_u\Bigl(
Z_{2:N+1}    
\Bigr)\\
&{\rm with}  \quad Z= 
\operatorname{Attn}\bigl(
        E_x(x_{t}),\,
        E_r(X^{R})
    \bigr)
\end{aligned}
\end{equation}
where $E_{x}:\mathbb{R}^{d_{\mathrm{state}}}\!\to\!\mathbb{R}^{d_{\mathrm{embed}}}$ encodes the current state, $E_{r}:\mathbb{R}^{N\times d_{\mathrm{ref}}}\!\to\!\mathbb{R}^{N\times d_{\mathrm{embed}}}$ embeds the reference horizon,  
\(\operatorname{Attn}(\cdot)\) denotes the multi-head self-attention stack, and  
\(D_{u}:\mathbb{R}^{d_{\mathrm{ffn}}}\!\rightarrow\!\mathbb{R}^{d_{\mathrm{action}}}\) is a nonlinear projection applied row-wise to produce the action sequence.

We deliberately favor an encoder-only architecture over an autoregressive decoder for three intertwined reasons.  
First, bidirectional self-attention is indispensable in finite-horizon MPC: every control input \(u_{t+i}\) must be selected with full awareness of all future reference tokens \(x^{R}_{t+j}\;(j\ge i)\).  
Encoder layers expose this global context at every depth, whereas the decoder’s causal mask blocks information flow from yet-to-be-generated tokens, inevitably degrading solution quality.  
Second, the encoder yields the entire action vector in a single, parallel pass, providing \(\mathcal{O}(1)\) inference latency with respect to the horizon length \(N\).  
By contrast, a decoder would require \(N\) sequential forward steps, accumulating delay and compounding numerical errors. Third, bidirectional attention permits the MPC loss to back-propagate simultaneously through every control input, ensuring stable optimization and coherent credit assignment across the horizon.  In a decoder, gradient signals would propagate step-by-step through the autoregressive chain, amplifying vanishing effects and slowing optimization. Taken together, these considerations make the encoder-only Transformer intrinsically aligned with the information flow and real-time constraints of horizon-aware MPC.

\subsection{Algorithm}
\label{subsec: Algorithm}

Having defined the policy structure, the final step is to determine parameters
$\theta^{\star}$ that satisfy~\eqref{eq.optimal approximation}.  In
principle, this can be achieved by either
(i) solution imitation or
(ii) policy gradient.  
The first paradigm solves the MPC problem off-line for a large set of
initial states and reference trajectories, then fits the policy network to the
resulting control sequences.  While conceptually simple, this procedure becomes prohibitive for
non-linear or high-dimensional plants because every optimal trajectory is
costly to compute and an extensive data set is required to avoid
out-of-distribution errors.  The second paradigm, adopted here, bypasses
optimal control sequence altogether and minimizes the true finite-horizon cost
\(
V(x_{t},X^{R},N;\theta)
\)
in~\eqref{eq:problem2} by gradient descent.  This removes the on-line
burden of repeatedly solving the MPC optimization and is therefore suitable
for complex dynamics. 

\begin{figure}[htbp]
    \centering
    \captionsetup{singlelinecheck = false, labelsep=period, font=small}
    \captionsetup[subfigure]{justification=centering}
    \includegraphics[width=0.5\textwidth]{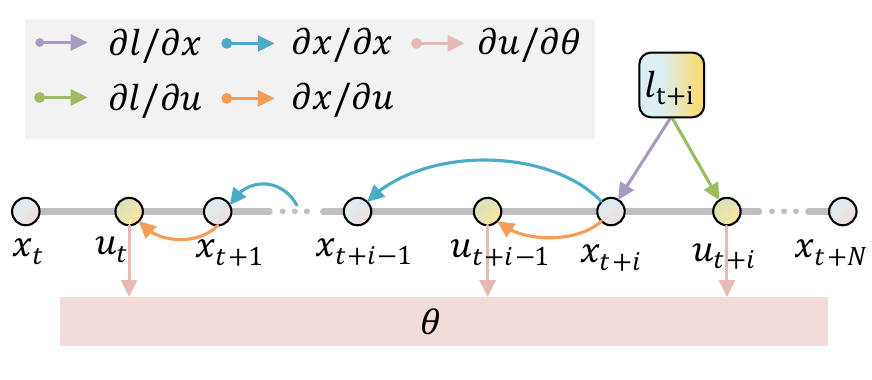}
    \caption{Gradient Propagation Process. Arrows show the gradient flow from the loss to the parameters.}
    \label{fig:grad_arch}
\end{figure}

We now detail TransMPC, which updates the Transformer policy by
directly minimizing the finite-horizon cost in~\eqref{eq:problem2}.  Let
$l_{t+i}=l\!\bigl(x_{t+i},x^{R}_{t+i},u_{t+i}\bigr)$ and \(u_{t+i}\) is the \(i\)-th control input of the control sequence
generated by \(\pi(x_{t},X^{R};\theta)\).   Differentiating
\eqref{eq:problem2} with respect to the policy parameters yields
\begin{equation}
\frac{{\rm d}V(x_t, X^R, N;\theta)}{{\rm d}\theta} = \sum_{i=1}^N \left[ \frac{\partial l_{t+i}}{\partial x_{t+i}} \frac{{\rm d} x_{t+i}}{{\rm d} \theta} + \frac{\partial l_{t+i}}{\partial u_{t+i}} \frac{{\rm d} u_{t+i}}{{\rm d} \theta} \right],
\label{eq:gradient}
\end{equation}
A key insight is that each control input depends on
$\theta$ explicitly through the Transformer policy, whereas every state
depends on $\theta$ only indirectly through the dynamics. Applying the chain rule to
$x_{t+1}=f(x_{t},u_{t})$ gives the recursive relation:
\begin{equation}
\frac{{\rm d} x_{t+i}}{{\rm d} \theta} = \frac{\partial f}{\partial x_{t+i-1}} \frac{{\rm d} x_{t+i-1}}{{\rm d} \theta} + \frac{\partial f}{\partial u_{t+i-1}} \frac{{\rm d} u_{t+i-1}}{{\rm d} \theta},
\label{eq:state_gradient}
\end{equation}
Note that $\frac{{\rm d} x_t}{{\rm d} \theta} = 0$ since the initial state is independent of the policy. The full gradient flow is visualized in
Fig.~\ref{fig:grad_arch}. 

To ensure that the policy generalizes across the entire state space and
accommodates variable prediction horizons, TransMPC minimises the
expected finite-horizon cost
\begin{equation}
\label{eq:batch_loss_final}
J(\theta)=
\mathbb{E}_{\substack{x\sim\mathcal{B}\\[2pt] N\sim\mathcal{U}[1,N_{\max}]}}
\bigl[\,V(x,X^{R},N;\theta)\bigr],
\end{equation}
where $\mathcal{B}$ is a replay buffer that stores states encountered
during exploration and $\mathcal{U}[1,N_{\max}]$ denotes the discrete
uniform distribution over horizon lengths.  Sampling $(x,N)$ in this way
provides i.i.d.\ minibatches and continuously exposes the network to a
range of sequence lengths, thereby promoting robust performance when the
horizon changes at run time.

\begin{figure*}[htbp]
    \centering
    \captionsetup{singlelinecheck = false, labelsep=period, font=small}
    \captionsetup[subfigure]{justification=centering}
    \includegraphics[width=1\textwidth]{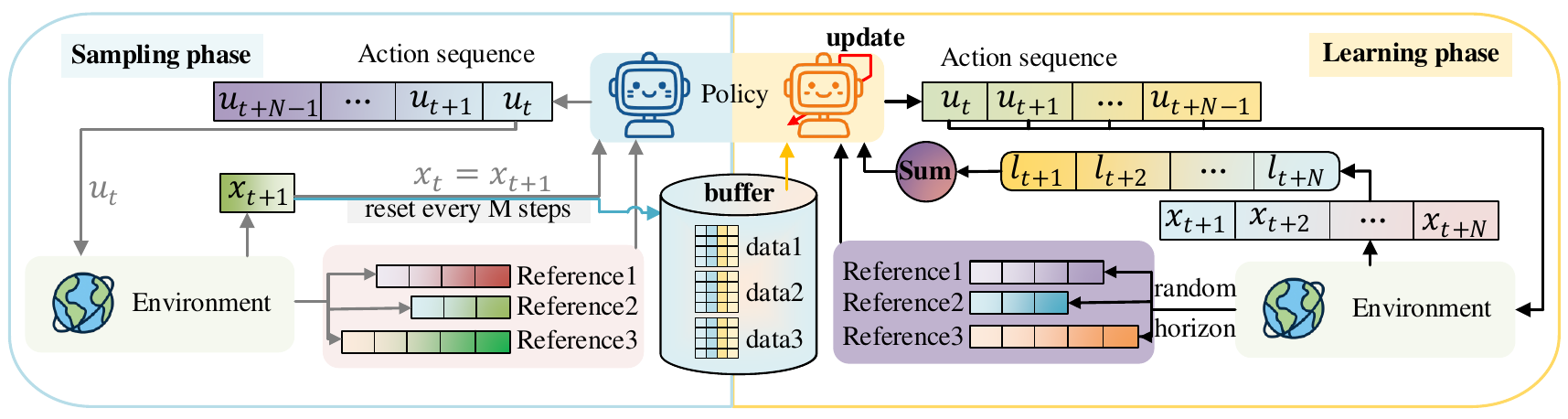}
    \caption{Sampling--learning training framework for $N$-step trajectory prediction. 
    In the \emph{sampling phase}, the agent executes an $N$-step action sequence, stores the resulting transitions in a replay buffer, and periodically resets the state. 
    In the \emph{learning phase}, the policy is updated by minimizing the cumulative loss over the entire prediction horizon.}
    \label{fig:train_arch}
\end{figure*}

Fig.~\ref{fig:train_arch} gives an overview of the training workflow,
which alternates between a sampling phase and a learning
phase.  In the sampling phase, the environment applies the first control input
$u_{t}$ generated by the current policy, stores the resulting transition
$(x_{t},u_{t},x_{t+1})$ in the replay buffer~$\mathcal{B}$.  To preserve state-space diversity, the interaction is truncated and the system is randomly reset every
$M$ steps.  In the learning phase, a minibatch of states is drawn from $\mathcal{B}$; for each state a horizon $N$ is sampled.  The full $N$-step action sequence generated by $\pi(\cdot;\theta)$ is rolled out through the model $f$ to compute $V$, after which the parameters are updated using the gradient derived in~\eqref{eq:gradient}.  Sampling and learning can proceed either
in lock-step or on separate threads, enabling concurrent data collection and optimization.

\begin{algorithm}
\caption{TransMPC}
\label{alg:transmpc}
\begin{algorithmic}[1]
\REQUIRE $\theta$: randomly initialized network parameter; $N_{\max}$: maximum horizon; $M$: episode length; $\alpha_\theta$: learning rate; $\mathcal{B}$: buffer
\STATE \textbf{while} not converged \textbf{do}
\STATE \quad // \textit{Sampling phase: Collect episodes with TransMPC, randomly reset every M steps:}
\STATE \quad \quad Sample horizon $N \sim \mathcal{U}[1,N_{\max}]$ \hfill $\triangleleft$ \textit{Sample horizon}
\STATE \quad \textbf{for} step $t = 0, \cdots, M-1$ \textbf{do}
\STATE \quad \quad  Calculate reference $X^R$ \hfill $\triangleleft$ \textit{Reference calculation}
\STATE \quad \quad $\mathbf{U} \leftarrow \pi(x_t, X^R; \theta)$ \hfill $\triangleleft$ \textit{Policy inference}
\STATE \quad \quad $u_t \leftarrow \mathbf{U}[0]$ \hfill $\triangleleft$ \textit{First action}
\STATE \quad \quad $x_{t+1} \sim f(x_t, u_t)$ \hfill $\triangleleft$ \textit{State Transition}
\STATE \quad \quad $\mathcal{B} \leftarrow \mathcal{B} \cup (x_{t+1})$ \hfill $\triangleleft$ \textit{Add to buffer}
\STATE \quad // \textit{Learning phase: Update TransMPC using collected data in $\mathcal{B}$:}

\STATE \quad \quad $x \sim \mathcal{B}$ and let $x_t=x$ \hfill $\triangleleft$ \textit{Sample minibatch}
\STATE \quad \quad Sample horizon $N \sim \mathcal{U}[1,N_{\max}]$ \hfill $\triangleleft$ \textit{Sample horizon}
\STATE \quad \quad Calculate reference $X^R$ \hfill $\triangleleft$ \textit{Reference generation}
\STATE \quad \quad $\mathbf{U} = \pi(x_t, X^R; \theta)$ \hfill $\triangleleft$ \textit{Generate action sequence}
\STATE \quad \quad $V = 0$ \hfill $\triangleleft$ \textit{Initialize cost accumulation}
\STATE \quad \quad \textbf{for} $i = 0..N-1$ \textbf{do}
\STATE \quad \quad \quad $V \leftarrow V + l(x_{t+i}, x^R_{t+i}, \mathbf{U}[i])$ \hfill $\triangleleft$ \textit{Accumulate cost}
\STATE \quad \quad \quad $x_{t+i+1} \leftarrow f(x_{t+i}, \mathbf{U}[i])$ \hfill $\triangleleft$ \textit{Model rollout}
\STATE \quad \quad Calculate $J(\theta) $ using \eqref{eq:batch_loss_final} 
\hfill $\triangleleft$  \textit{Loss calculation} 
\STATE \quad \quad $\theta \leftarrow \theta - \alpha_\theta \frac{{\rm d}J(\theta)}{{\rm d} \theta} $ 
\hfill $\triangleleft$ \textit{Policy update}
\end{algorithmic}
\end{algorithm}

\section{Simulation Validation}
\label{sec: simulation validation}
The effectiveness of TransMPC is assessed on a combined
longitudinal–lateral vehicle-tracking task, a well-established benchmark
for nonlinear, non-affine MPC problems
\cite{li2017driver,duan2021adaptive,cheng2020model,ji2016path}.

\subsection{Task Description}
\label{subsec: task description}

We adopt the nonlinear bicycle model of Ge \textit{et al.}~\cite{ge2021numerically}; the complete state and input definitions are listed in Table~\ref{tab:state_and_input}.  
Actuator limits are imposed as 
\(a_{x}\in[-3,\,3]~\text{m/s}^{2}\) for longitudinal acceleration and 
\(\delta\in[-0.52,\,0.52]~\text{rad}\) for steering.  
Two reference paths, including a sinusoidal lane and a double‐lane change, are tracked at a nominal speed of \(5~\text{m/s}\).  
The running cost is
\begin{equation}
\label{eq.reward_function_tracking}
\begin{aligned}
l_{\rm track} = & 0.2(p_x - p_x^{\mathrm{R}})^2 + 0.3(p_y - p_y^{\mathrm{R}})^2 \\
& + 0.2(\varphi - \varphi^{\mathrm{R}})^2 + 0.3(v - v^{\mathrm{R}})^2 \\
& + 0.1 v^2 + 0.1 \omega^2 + 0.05 a_x^2 + 0.05 \delta^2,
\end{aligned}
\end{equation}
where \((p_x^{\mathrm R}, p_y^{\mathrm R}), \varphi^{\mathrm R}\), and
\(v^{\mathrm R}\) denote the reference position, yaw, and speed,
respectively.  
This function penalizes tracking error, and regularizes speed, yaw rate, and control effort.

\begin{table}[!htb]
\captionsetup{justification=centering,labelsep=newline,font={small}}
    \centering
    \caption{State, reference, and control inputs}
    \label{tab:state_and_input}
    \begin{tabular}{c l c c}
    \toprule
    Category & Symbol & Description & Unit \\
    \midrule
    \multirow{6}{*}{States}
    & $p_x$ & Longitudinal position & m \\
    & $p_y$ & Lateral position & m \\
    & $\varphi$ & Heading angle & rad \\
    & $v$ & Longitudinal velocity & m/s \\
    & $u$ & Lateral velocity & m/s \\
    & $\omega$ & Yaw rate at CoG & rad/s \\
    \midrule
    \multirow{4}{*}{Reference}
    & $p_x^\mathrm{R}$ & Reference longitudinal position & m \\
    & $p_y^\mathrm{R}$ & Reference lateral position & m \\
    & $\varphi^\mathrm{R}$ & Reference heading angle & rad \\
    & $v^\mathrm{R}$ & Reference velocity & m/s \\
    \midrule
    \multirow{2}{*}{Inputs}
    & $a_x$ & Longitudinal acceleration & m/s\textsuperscript{2} \\
    & $\delta$ & Steering angle & rad \\
    \bottomrule
    \end{tabular}
\end{table}

\subsection{Algorithm Details} 
\label{subsec: algorithm details}

\begin{figure}[htbp]
    \centering
    \captionsetup{singlelinecheck=false, labelsep=period, font=small}
    \captionsetup[subfigure]{justification=centering}
    \includegraphics[width=0.5\textwidth]{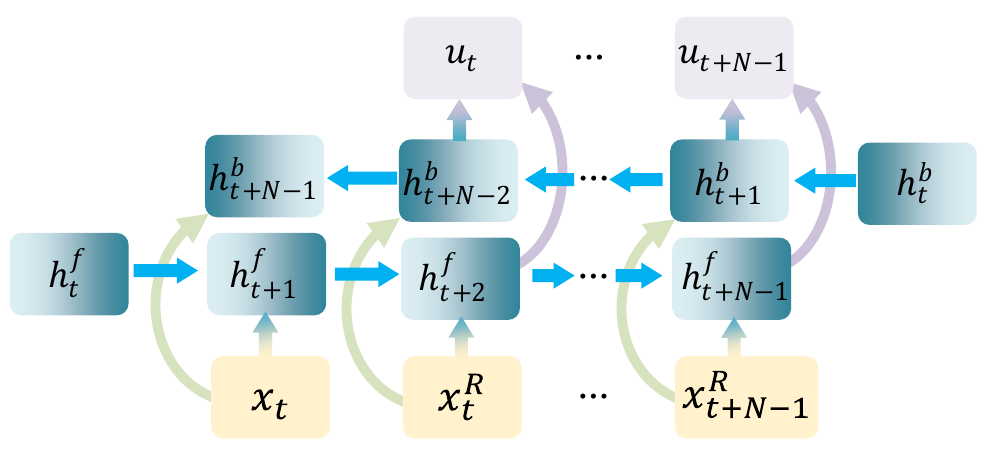}
    \caption{Bidirectional hidden-state loop for optimal action sequencing. 
    The loop fuses forward (\(h_{t-1}^{f}\)) and backward (\(h_{t-1}^{b}\)) hidden states at each time step~\(t\) to generate an MPC-compatible control sequence \(u_{t},\dots,u_{t+N}\).}
    \label{fig:bi-directional}
\end{figure}

TransMPC is evaluated against four baselines.  
MPC-MLP (identical to the FHADP \cite{li2023reinforcement} scheme) employs an MLP to represent the explicit policy. The original RMPC \cite{liu2022recurrent} employs a unidirectional GRU \cite{8053243} and therefore emits only a
single control input per step; we replace the GRU with its bidirectional
counterpart to obtain biRMPC. We likewise convert the recently introduced Mamba and TTT recurrent
architectures \cite{gu2023mamba,sun2020test} into bidirectional form,
creating the baselines MPC--Mamba and MPC--TTT.
Fig.~\ref{fig:bi-directional} illustrates the shared bidirectional
design: the hidden state at time~\(t\) aggregates information from both
past and future reference tokens, ensuring that every control output
\(u_{t+i}\) is conditioned on the complete horizon \(X^{R}\).  

To guarantee fairness in evaluation, all methods share identical baseline hyperparameters: a model embedding dimension \(d_{\text{embed}}\) of 256, a learning rate of \(10^{-3}\) using the Adam optimizer. Despite architectural distinctions, all compared methods maintain similar parameter counts and computational complexity, ensuring performance disparities reflect inherent differences in modeling capability rather than variations in model capacity. Specifically, the TransMPC employs a 4-head attention mechanism with the feedforward dimension consistent with the \(d_{\text{embed}}\). The key parameters of MPC--Mamba include a state dimension of 32 and a convolution dimension of 4, which control its state space memory capacity and local receptive field size. MPC--TTT utilizes a convolution dimension of 4, with its core test-time learning rate aligned with the baseline learning rate. MPC--biGRU maintains a hidden state dimension of 256. Lastly, MPC--MLP is implemented as a 3-layer multilayer perceptron network structure, with each layer having a dimension of 256. The maximum prediction horizon is set to 20.

\subsection{Results}

We evaluate the performance of TransMPC in terms of solution accuracy, closed-loop tracking quality, and computational efficiency. Solution accuracy is assessed using 200 randomly sampled initial states based on the normalized error defined as:
\[
\text{Relative Accuracy} = \frac{\lvert u^{\theta}-u^{\star}\rvert}{u_{\max}-u_{\min}},
\]
where \(u^{\theta}\) represents the control input generated by the evaluated method, \(u^{\star}\) is the optimal solution obtained via IPOPT-based MPC solver \cite{Andreas2006Biegler}, and \(u_{\max}\), \(u_{\min}\) denote the global maximum and minimum bounds of the control input, respectively. For multi-dimensional control inputs, this accuracy metric is calculated element-wise and reported individually for each control dimension.

\begin{figure}[t]
\centering
\captionsetup{singlelinecheck = false,labelsep=period, font=small}
\captionsetup[subfigure]{justification=centering}
    \subfloat[\label{fig:option}]
    {\includegraphics[width = 0.5\textwidth]{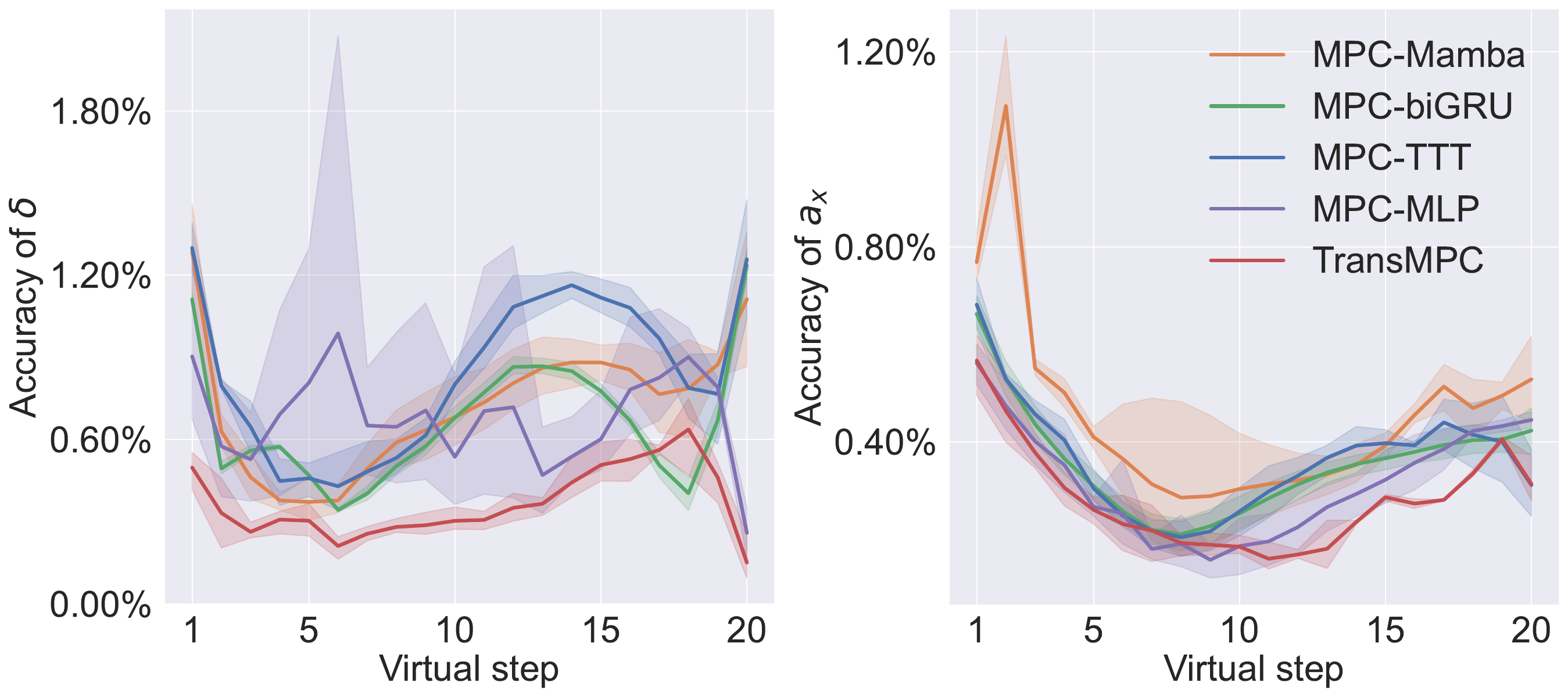}} \quad
    \subfloat[\label{fig:option2}]
    {\includegraphics[width = 0.5\textwidth]{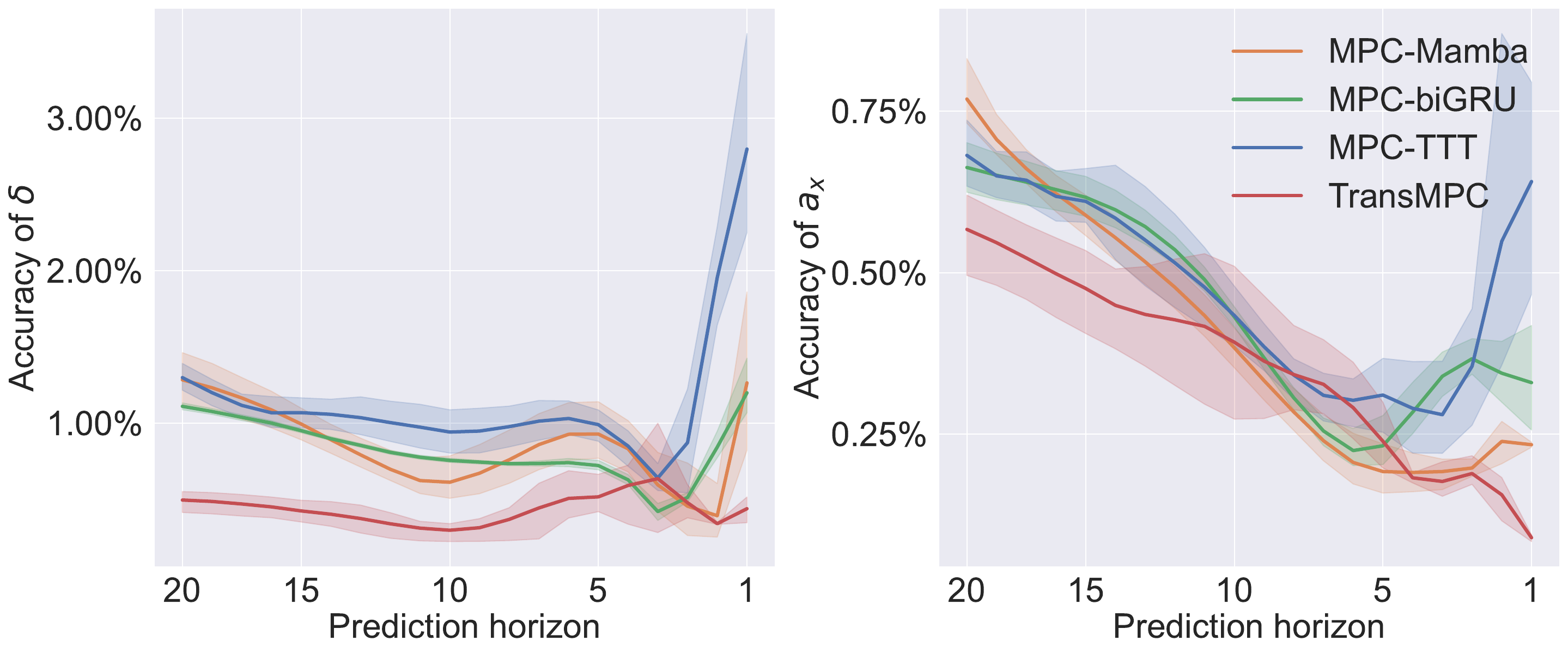}} 
    \caption{Relative accuracy of TransMPC control inputs. (a) Element-wise accuracy of the entire control sequence when \(N = 20\).  (b) Accuracy of the first control input \(u_t\) as the prediction horizon varies from 1 to 20.}
    \label{fig:option1}
\end{figure}

Fig.~\ref{fig:option1} provides a comparison of the overall solution accuracy. Fig.~\ref{fig:option} further details the accuracy across the entire control sequence at a fixed prediction horizon of \(N=20\). TransMPC consistently demonstrates higher accuracy than other methods throughout the prediction horizon, closely approximating the optimal MPC solution. Among the predicted control inputs, the first element \(u_t\) is particularly critical, as it is directly executed in practice. Thus, Fig.~\ref{fig:option2} specifically evaluates the accuracy of \(u_t\) for horizons varying from 1 to 20. TransMPC maintains superior accuracy across all tested horizons, demonstrating robust adaptability to different prediction lengths.

\begin{figure}[t]
    \centering
    \captionsetup{singlelinecheck = false,labelsep=period, font=small}
    \captionsetup[subfigure]{justification=centering}
    \begin{minipage}{0.45\textwidth}
        \centering
        \includegraphics[width=\textwidth]{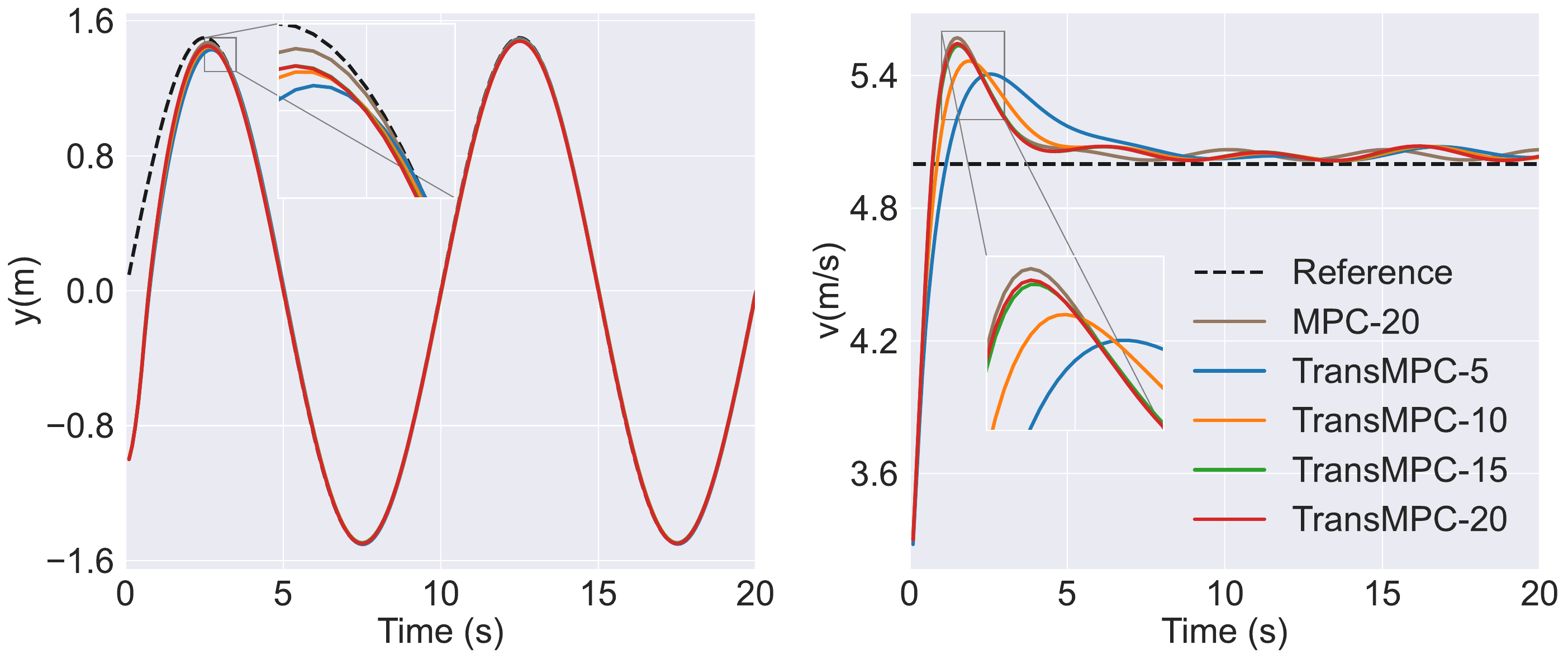}
    \end{minipage}\hfill
    \begin{minipage}{0.45\textwidth}
        \centering
        \subfloat[Sin curve trajectory tracking\label{subFig:State curve0}]{\includegraphics[width=\textwidth]{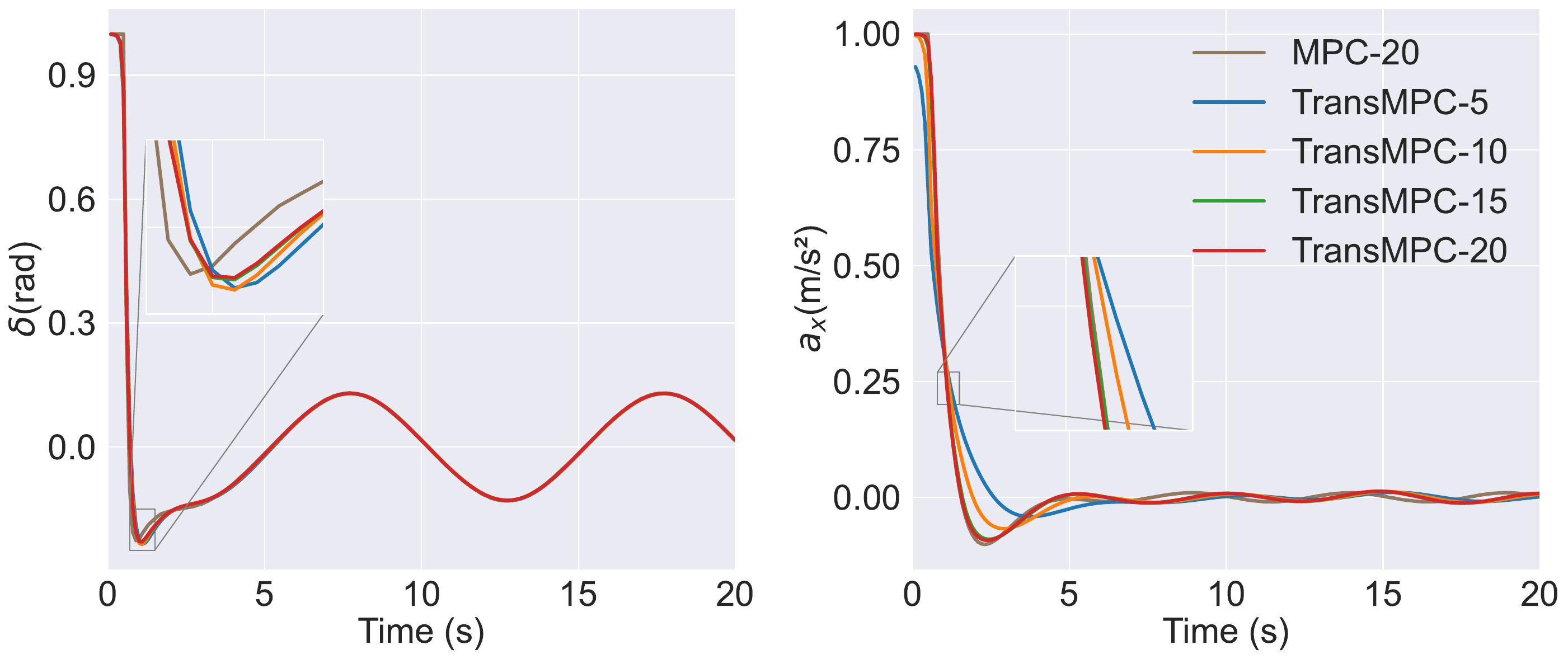}}
    \end{minipage}\\
    \begin{minipage}{0.45\textwidth}
        \centering
        \includegraphics[width=\textwidth]{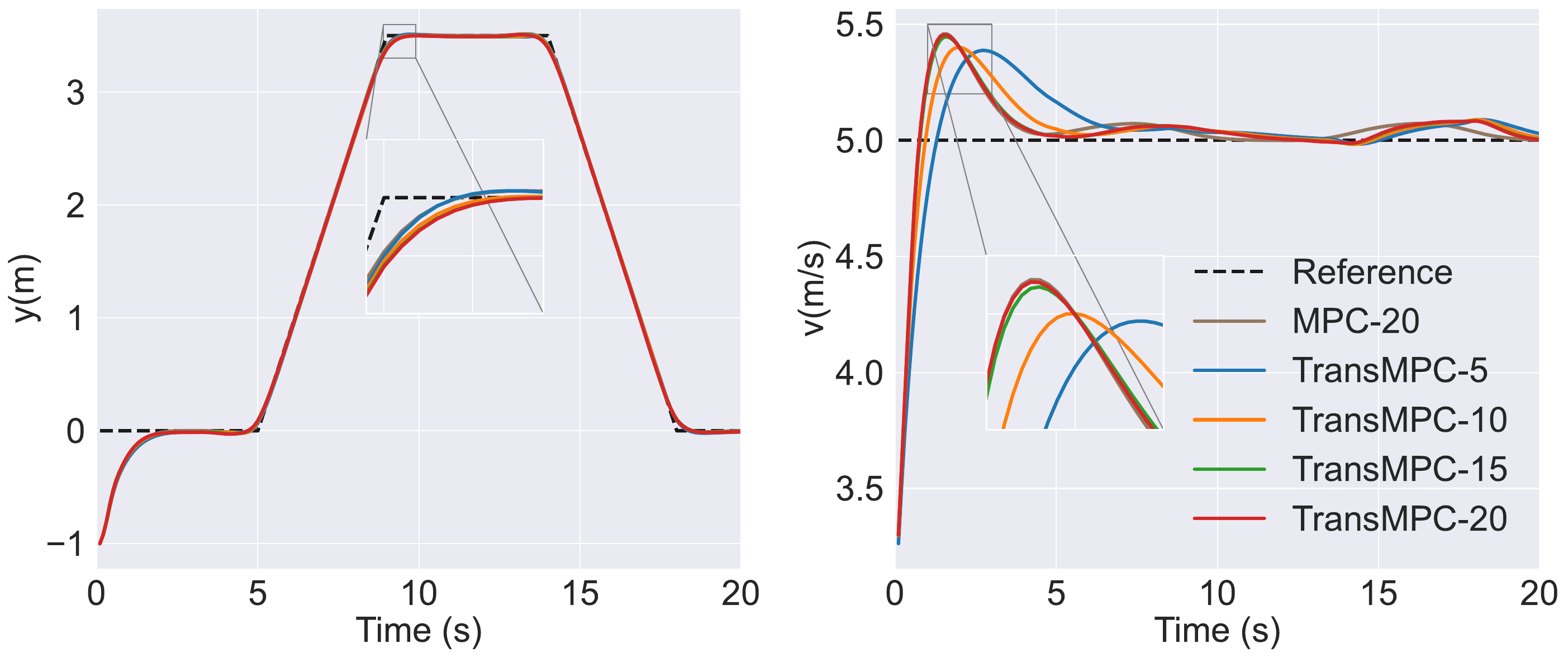}
    \end{minipage}\hfill
    \begin{minipage}{0.45\textwidth}
        \centering
        \subfloat[Double lane shift trajectory tracking\label{subFig:State curve1}]{\includegraphics[width=\textwidth]{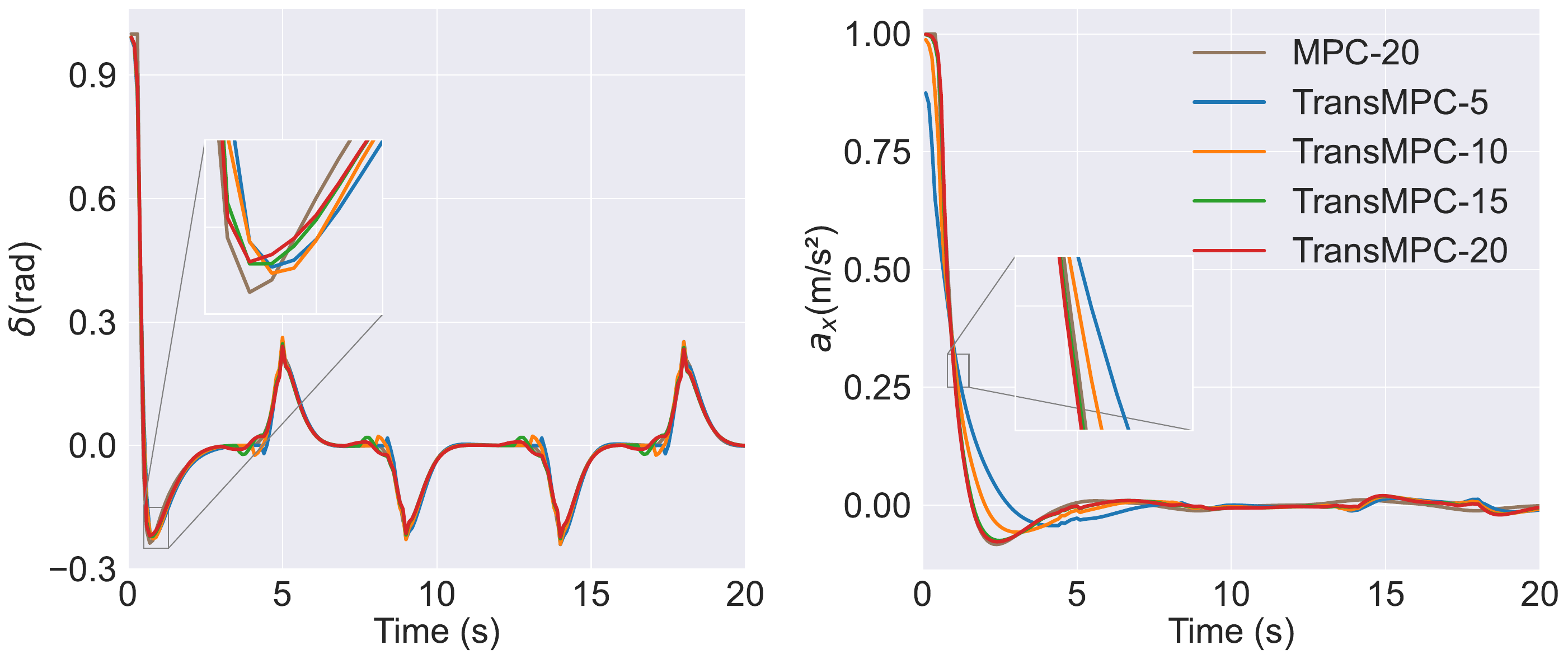}}
    \end{minipage}
    \caption{State and control curves of TransMPC under different prediction horizons.}
    \label{fig: State curve}
\end{figure}

Fig.~\ref{fig: State curve} shows representative tracking results for sinusoidal and double-lane-change trajectories controlled by TransMPC with different prediction horizons. TransMPC demonstrates consistent and accurate tracking performance across various horizon lengths. Quantitative results are further summarized in Tables~\ref{tab:prediction_horizon} and~\ref{tab:prediction_horizon_dlc}, presenting average tracking errors and cumulative control costs ($C$) computed over the entire control period (170 control steps at a frequency of 10Hz). Among these algorithms, MPC-Mamba, MPC-biGRU, MPC-TTT, and TransMPC inherently support variable prediction horizons, while MPC-MLP requires retraining when the horizon changes. As anticipated, extending the prediction horizon generally results in improved tracking performance. Results indicate that TransMPC consistently achieves performance closest to the optimal MPC solution in both the sinusoidal and double-lane-change scenarios. Interestingly, when employing a short prediction horizon of 5 steps, TransMPC unexpectedly surpasses MPC in performance, which contradicts the common intuition since MPC serves as the theoretical ground truth. This phenomenon occurs because TransMPC implicitly learns to infer and leverage reference information extending beyond the immediate five-step prediction horizon, thereby enhancing its overall predictive capability.

\begin{table}[!htb]
\captionsetup{justification=centering, labelsep=newline, font={small}}
\centering
\caption{Performance comparison for the sinusoidal scenario}
\label{tab:prediction_horizon}
\begin{threeparttable}
\begin{tabular}{c@{\hspace{2pt}}c@{\hspace{6pt}}*{6}{c@{\hspace{2pt}}}}
\toprule
\multirow{2}{*}{$N$} & \multirow{2}{*}{Metrics} & \multicolumn{6}{c}{Algorithms} \\ \cmidrule(l){3-8}
 &  & \scriptsize TransMPC & \scriptsize MPC-Mamba & \scriptsize MPC-biGRU & \scriptsize MPC-TTT & \scriptsize MPC-MLP & \scriptsize MPC \\ \midrule
\multirow{2}{*}{20}
 & $\Delta y\,[\mathrm{m}]$ & \textbf{0.008} & 0.020 & 0.015 & 0.013 & 0.013 & 0.005 \\
 & $C$                      & \textbf{0.043} & 0.051 & 0.051 & 0.048 & 0.049 & 0.034 \\ \midrule
\multirow{2}{*}{15}
 & $\Delta y\,[\mathrm{m}]$ & \textbf{0.009} & 0.022 & 0.015 & 0.014 & 0.014 & 0.008 \\
 & $C$                      & \textbf{0.048} & 0.052 & 0.051 & 0.060 & 0.050 & 0.039 \\ \midrule
\multirow{2}{*}{10}
 & $\Delta y\,[\mathrm{m}]$ & \textbf{0.012} & 0.023 & 0.015 & 0.014 & 0.016 & 0.013 \\
 & $C$                      & \textbf{0.083} & 0.105 & 0.099 & 0.106 & 0.096 & 0.070 \\ \midrule
\multirow{2}{*}{5}
 & $\Delta y\,[\mathrm{m}]$ & \textbf{0.015} & 0.052 & 0.035 & 0.049 & 0.060 & 0.032 \\
 & $C$                      & \textbf{0.251} & 1.278 & 0.696 & 2.482 & 3.194 & 1.214 \\
\bottomrule
\end{tabular}
\end{threeparttable}
\end{table}

\begin{table}[!htb]
\captionsetup{justification=centering, labelsep=newline, font={small}}
\centering
\caption{Performance comparison for the double lane change}
\label{tab:prediction_horizon_dlc}
\begin{threeparttable}
\begin{tabular}{c@{\hspace{2pt}}c@{\hspace{6pt}}*{6}{c@{\hspace{2pt}}}}
\toprule
\multirow{2}{*}{$N$} & \multirow{2}{*}{Metrics} & \multicolumn{6}{c}{Algorithms} \\ \cmidrule(l){3-8}
 &  & \scriptsize TransMPC & \scriptsize MPC-Mamba & \scriptsize MPC-biGRU & \scriptsize MPC-TTT & \scriptsize MPC-MLP & \scriptsize MPC \\ \midrule
\multirow{2}{*}{20}
 & $\Delta y\,[\mathrm{m}]$ & \textbf{0.015} & 0.022 & 0.022 & 0.021 & 0.015 & 0.011 \\
 & $C$                      & \textbf{0.041} & 0.048 & 0.050 & 0.052 & 0.045 & 0.033 \\ \midrule
\multirow{2}{*}{15}
 & $\Delta y\,[\mathrm{m}]$ & \textbf{0.017} & 0.023 & 0.022 & 0.021 & 0.017 & 0.011 \\
 & $C$                      & \textbf{0.041} & 0.052 & 0.059 & 0.052 & 0.049 & 0.037 \\ \midrule
\multirow{2}{*}{10}
 & $\Delta y\,[\mathrm{m}]$ & \textbf{0.017} & 0.025 & 0.023 & 0.022 & 0.019 & 0.014 \\
 & $C$                      & \textbf{0.081} & 0.093 & 0.089 & 0.083 & 0.096 & 0.064 \\ \midrule
\multirow{2}{*}{5}
 & $\Delta y\,[\mathrm{m}]$ & \textbf{0.018} & 0.037 & 0.025 & 0.034 & 0.040 & 0.023 \\
 & $C$                      & \textbf{0.223} & 1.037 & 0.617 & 1.745 & 2.750 & 1.036 \\
\bottomrule
\end{tabular}
\end{threeparttable}
\end{table}

\begin{figure}
\centering
\captionsetup{singlelinecheck = false,labelsep=period, font=small}
\captionsetup[subfigure]{justification=centering}
\includegraphics[width=0.3\textwidth]{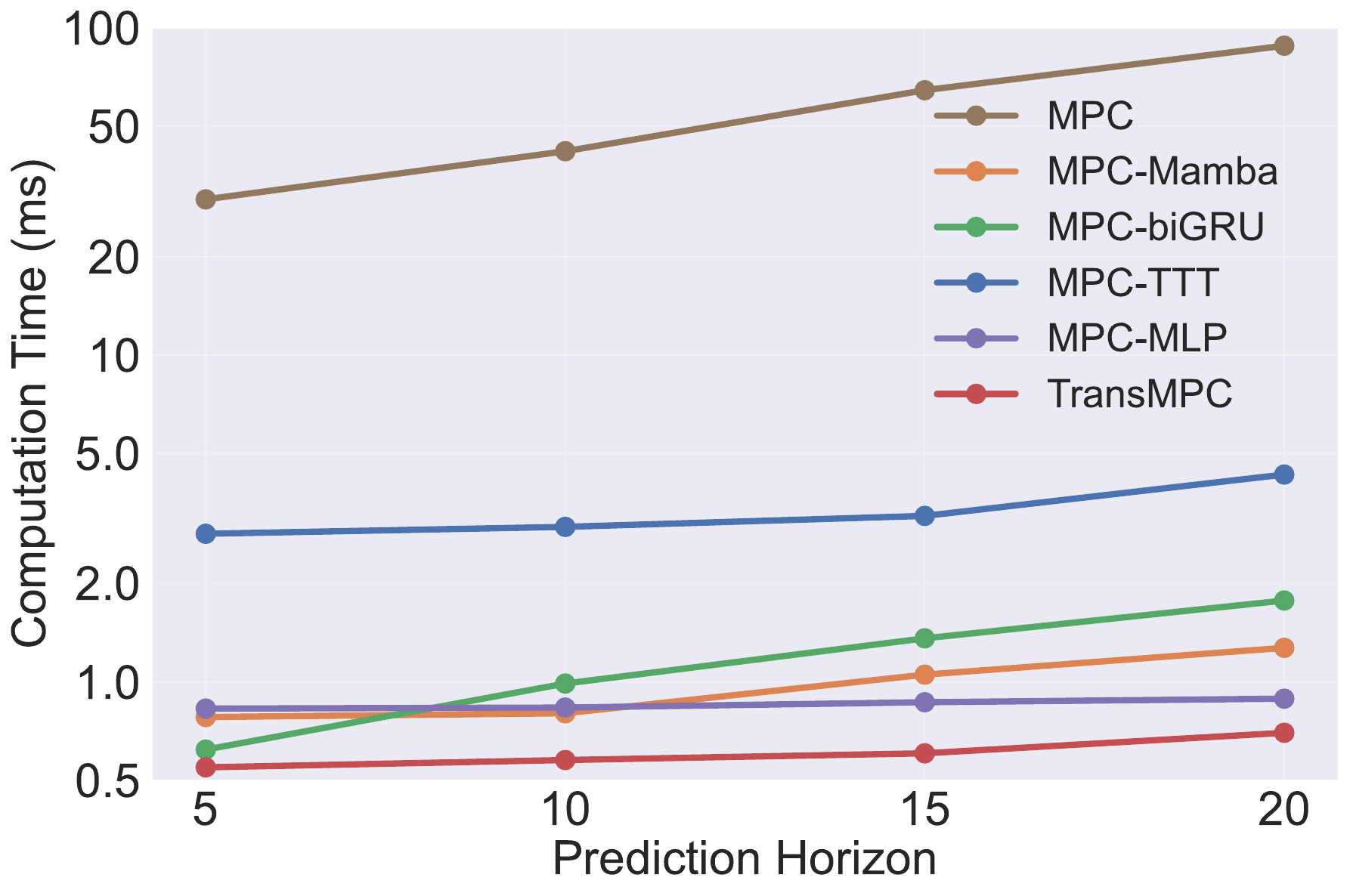}
\caption{Computation time per control sequence at different prediction horizons.
}
\label{fig. compute_time}
\end{figure}

Computation time per control sequence was evaluated starting from the second inference step, as MPC-Mamba requires initialization of hidden states at the first step. Fig.~\ref{fig. compute_time} demonstrates that TransMPC consistently outperforms the other methods across all tested horizons. For a prediction horizon of 20, TransMPC achieves computational speeds that are 81.97\% faster than MPC-Mamba, 153.93\% faster than MPC-biGRU, 516.74\% faster than MPC-TTT, and 27.18\% faster than MPC-MLP. Notably, unlike competing approaches, TransMPC maintains stable computation efficiency regardless of horizon length and does not require retraining, making it highly suitable for practical applications.

\section{Experimental Validation}
\label{sec: experimental validation}
In this section, we validate the real-world performance of TransMPC on a trajectory tracking task with obstacle avoidance.

\subsection{Task Description}
\label{subsec: Task description}
We conduct experiments using an Autolabor mobile robot, which is a logistics robot with differential steering. Key parameters of the vehicle are summarized in Table~\ref{tab. Vehicle parameters}. The control inputs are defined as incremental linear velocity \(\delta v\) (m/s\textsuperscript{2}) and incremental angular velocity \(\delta \omega\) (rad/s). The robot’s kinematic model is expressed as follows:
\begin{equation}
\begin{aligned}
x_{t+1} = x_t +
\begin{bmatrix}0 \\0 \\0 \\ \delta{v} \\\delta{\omega} \end{bmatrix} + 
\begin{bmatrix}v_t \cos \varphi_{t} \\v_t \sin \varphi_{t} \\ \omega_t \\ 0 \\ 0 \end{bmatrix} /f,
\end{aligned}
\end{equation}
where \(f = 10\) Hz is the control frequency, and control increments are bounded by \(\lvert \delta v \rvert \leq 0.8/f\) and \(\lvert \delta \omega \rvert \leq 0.4/f\).

\begin{figure}
\centering
\captionsetup{singlelinecheck = false,labelsep=period, font=small}
\captionsetup[subfigure]{justification=centering}
\includegraphics[width=0.45\textwidth]{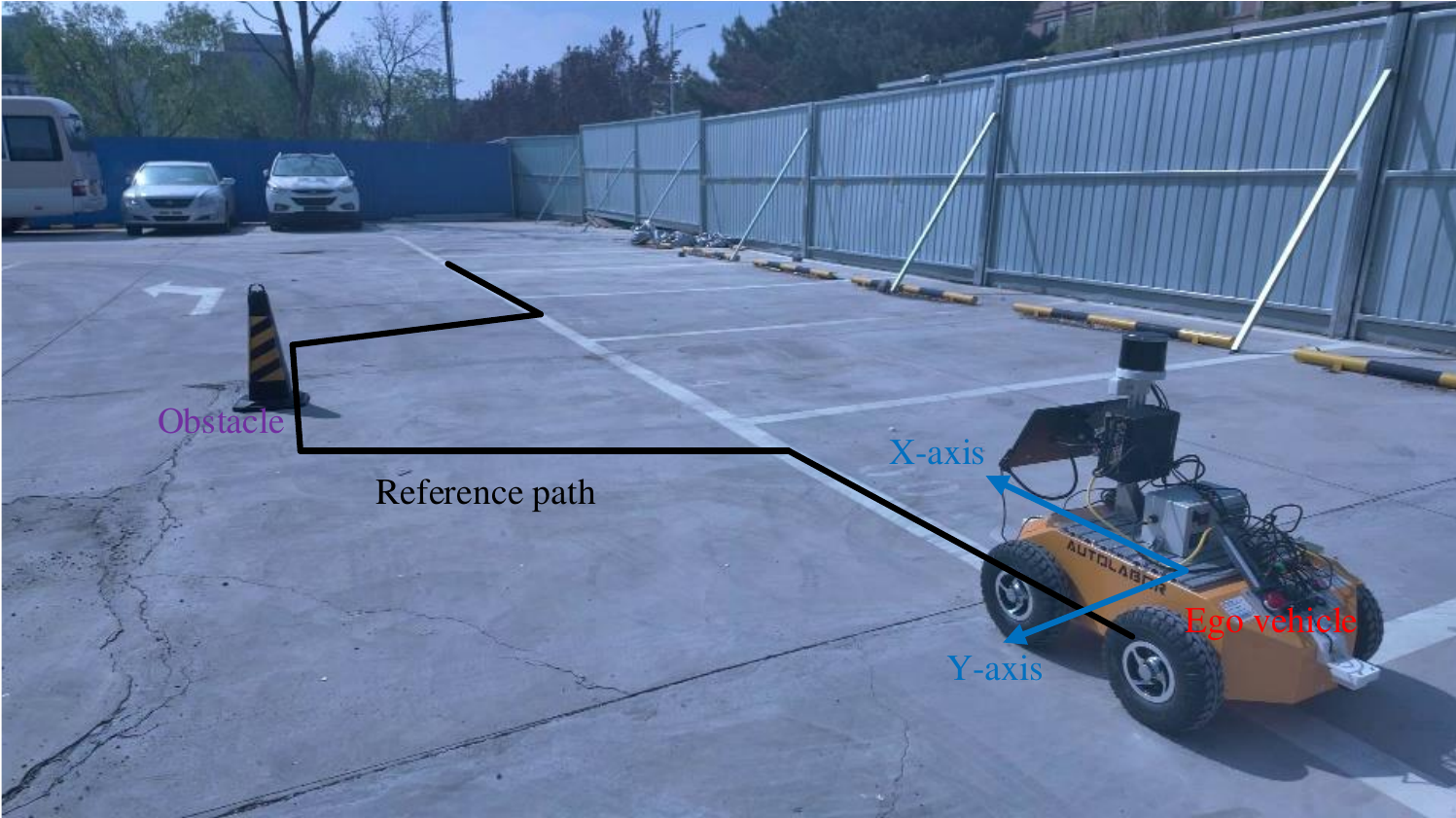}
\caption{Vehicle trajectory tracking and collision avoidance task.}
\label{fig. vehicle tracking task}
\end{figure}

\begin{figure}
\centering
\captionsetup{singlelinecheck = false,labelsep=period, font=small}
\captionsetup[subfigure]{justification=centering}
\includegraphics[width=0.48\textwidth]{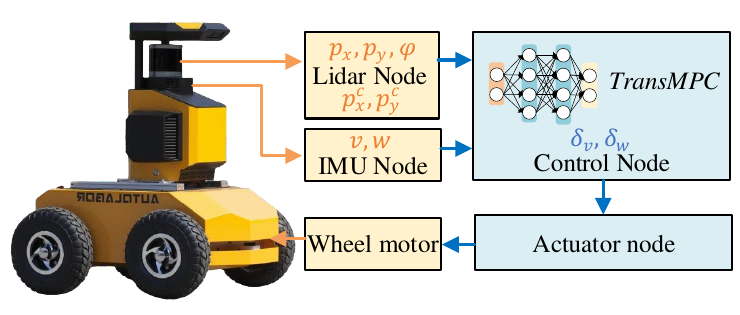}
\caption{Experiment framework.}
\label{fig. ros framework}
\end{figure}

The tracking and obstacle avoidance scenario is illustrated in Fig.~\ref{fig. vehicle tracking task}. The ego vehicle must follow a given reference trajectory at a desired speed of \(0.4\,\text{m/s}\), while simultaneously avoiding obstacles randomly positioned along the reference path. The vehicle employs a LiDAR and an IMU for localization, velocity, and angular velocity measurement. Additionally, obstacle positions are identified through LiDAR scans. The trained policy generates control inputs based on sensor data, which are transmitted via ROS to the robot's actuator nodes, forming a complete closed-loop control system as depicted in Fig.~\ref{fig. ros framework}.

\begin{table}[!htb] \captionsetup{justification=centering,labelsep=newline,font=small}
\captionsetup{justification=centering,labelsep=newline,font={small}}
      \centering
	\caption{Vehicle parameters}
	\centering
	\label{tab. Vehicle parameters}
		\begin{tabular}{cccc}
			\toprule
			Description & Symbol & Value\\ \cmidrule(l){1-3} 
% boundary dimension & $\mathrm{length} \times \mathrm{width} \times 
rotation diameter & $d$ & 953 mm \\
wheel track & $d_\mathrm{w}$ & 523 mm\\
Tyre diameter & $D$ & 260 mm\\
mass & $m$ & 50 kg \\
size & $l_x \times l_y \times l_z$  & $726 \times 617 \times 273  \mathrm{mm}$ \\
			\bottomrule
		\end{tabular}
\end{table}

Compared to the pure trajectory tracking task described in Section~\ref{sec: simulation validation}, the running cost in this task includes an additional collision avoidance term:
\begin{equation}
\label{eq.Tracking_collision_avoidance_reward}
\begin{aligned}
l &= l_{\rm track} - {l_{\mathrm{c}}}^2 \\
l_{\mathrm{c}} &= \sqrt{(p_x^c-p_x)^2 + (p_y^c-p_y)^2} - (r_{\text{ego}} + r_{\text{obstacle}} + r_{\text{safe}})
\end{aligned}
\end{equation}
where \(l_{\mathrm{c}}\) represents collision proximity, \((p_x^c, p_y^c)\) are the coordinates of the obstacle, \(r_{\text{ego}}\) and \(r_{\text{obstacle}}\) denote radii of the ego vehicle and obstacle respectively, and \(r_{\text{safe}}=0.1\,\text{m}\) specifies the additional safety margin. The full state vector is defined as:
\begin{equation}
\label{eq.tracking_collision_avoidance_obs}
x = \{p_x, p_y, \varphi, v, \omega, p_x^c - p_x, p_y^c - p_y\},
\end{equation}
combining vehicle state information and relative obstacle positions. 

\subsection{Results}
\begin{figure}[t]
    \centering
    \captionsetup{singlelinecheck = false,labelsep=period,font=small}
    \captionsetup[subfigure]{justification=centering}
    
    \begin{minipage}{0.48\textwidth}
        \centering
        \subfloat[\label{subFig:vehicle_trajectory}]
                 {\includegraphics[width=\textwidth]{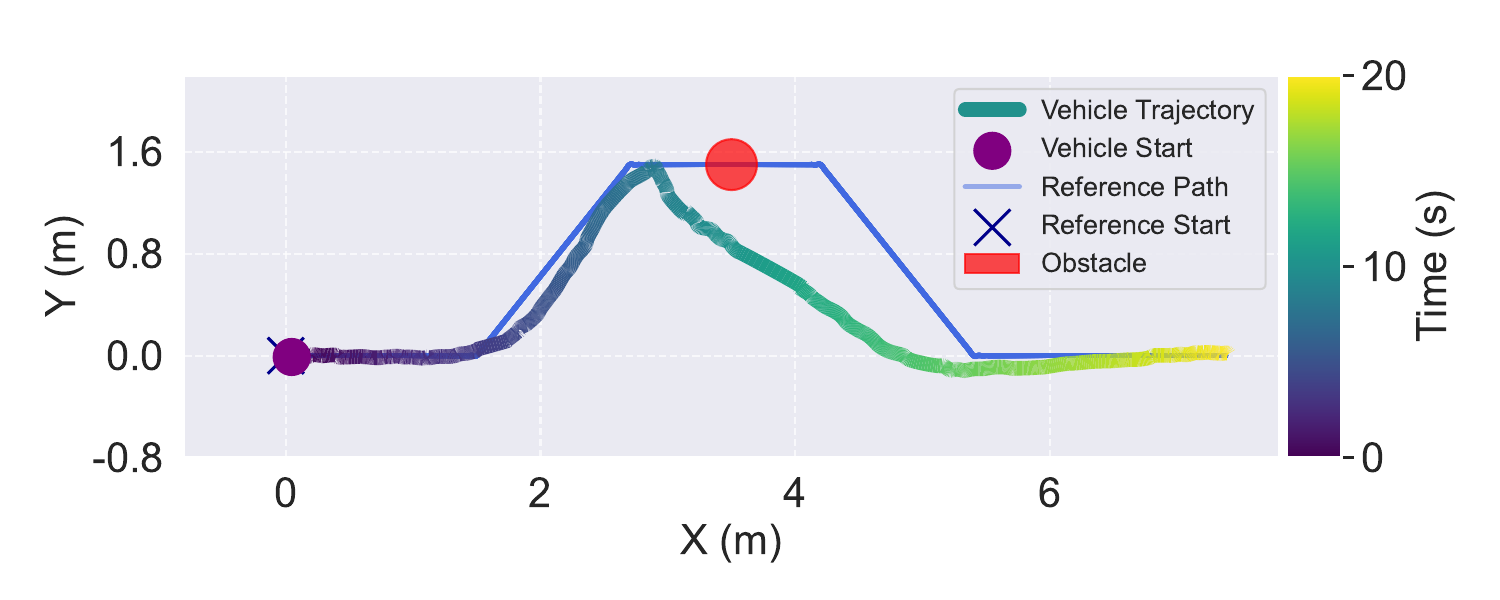}}
    \end{minipage}\\
    
    \begin{minipage}{0.24\textwidth}
        \centering
        \subfloat[\label{subFig:velocity_command}]
                 {\includegraphics[width=\textwidth]{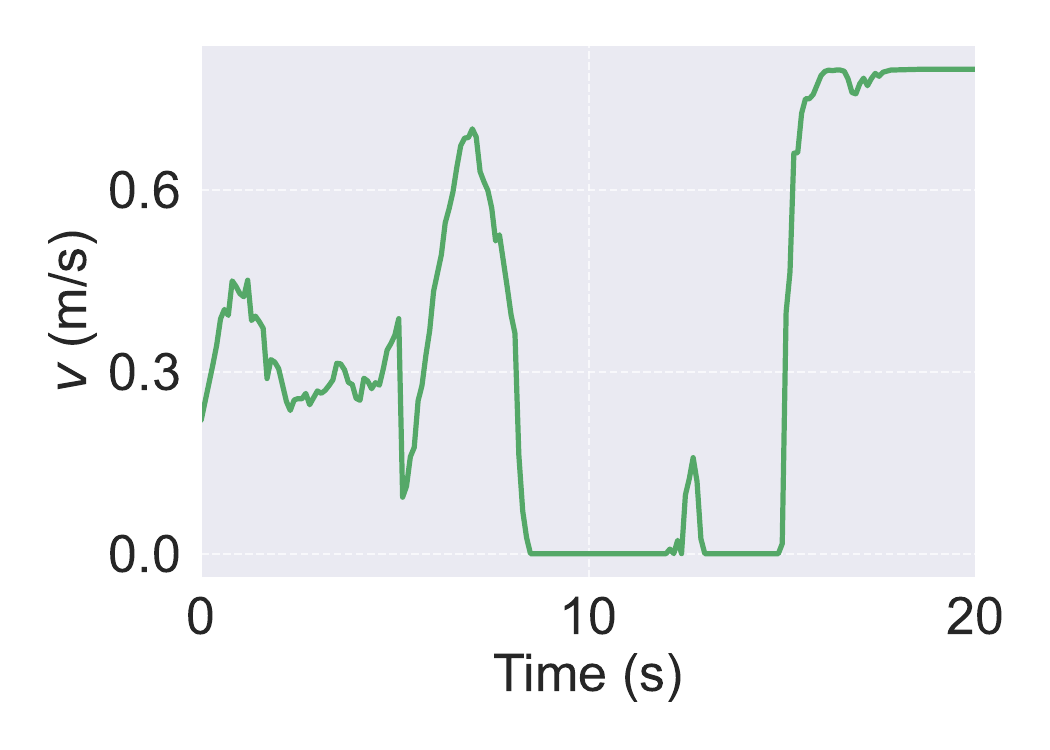}}
    \end{minipage}\hfill
    \begin{minipage}{0.24\textwidth}
        \centering
        \subfloat[\label{subFig:angular_velocity_command}]
                 {\includegraphics[width=\textwidth]{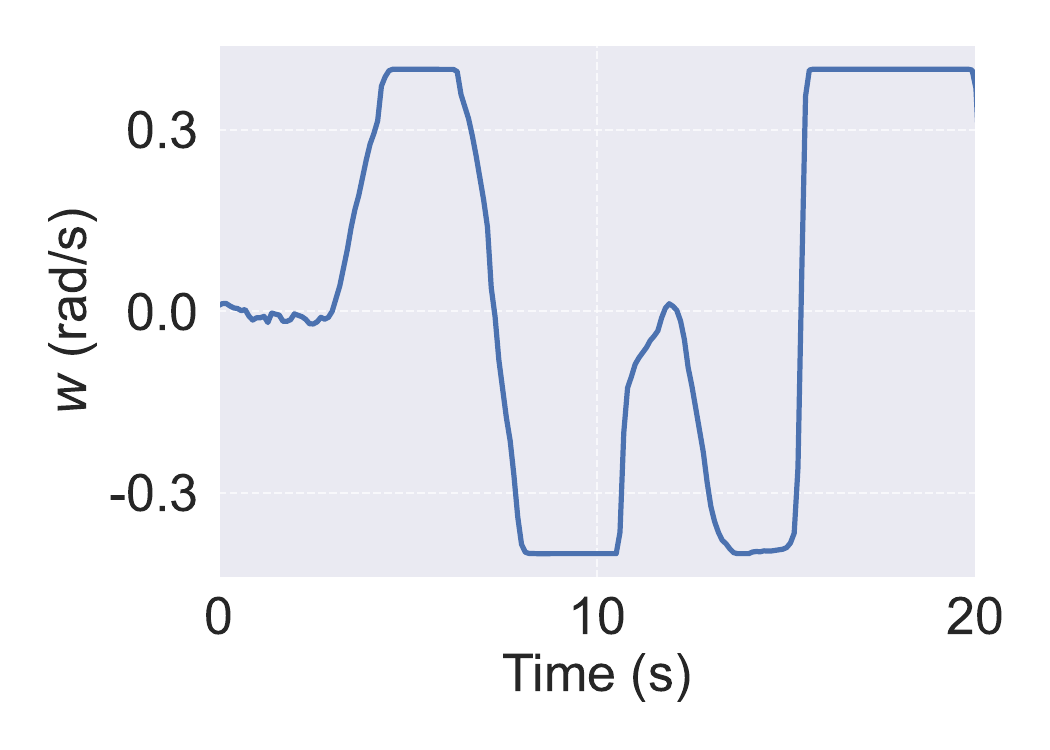}}
    \end{minipage}
    
    \caption{Experimental results. (a) Vehicle trajectory. (b) Velocity curves. (c) Angular velocity curves.}
    \label{fig:vehicle_analysis}
\end{figure}

\begin{figure}
\centering
\captionsetup{singlelinecheck = false,labelsep=period, font=small}
\captionsetup[subfigure]{justification=centering}
\includegraphics[width=0.45\textwidth]{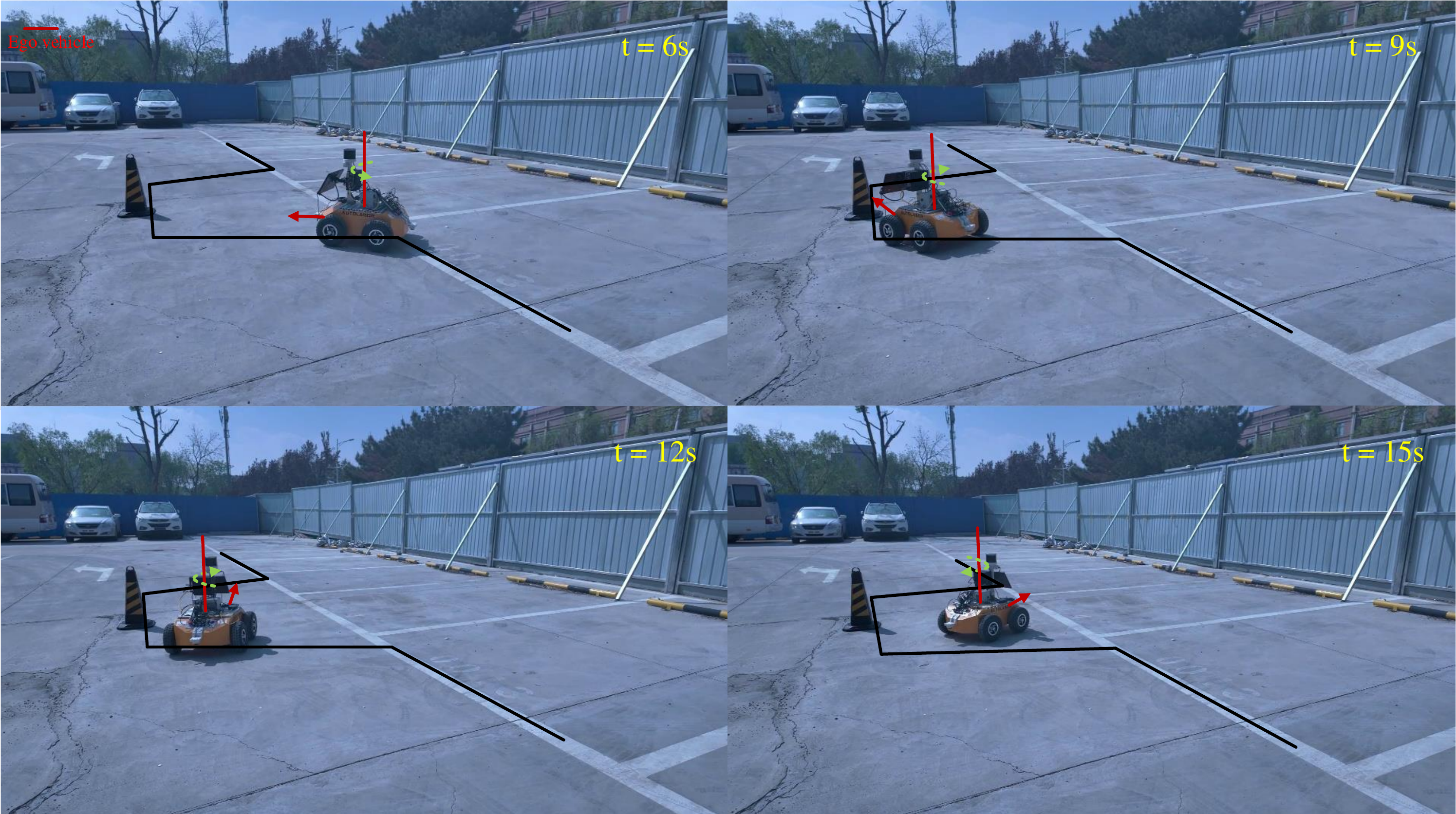}
\caption{Experimental snapshots. The red arrow marks the movement direction, while the yellow arrow indicates its rotation. At $t=9$s, the ego-vehicle encounters an obstacle and maintains a generous clearance. By $t=12$s, it has safely driven past the obstacle.}
\label{fig. vehicle experiments}
\end{figure}

The TransMPC policy was trained with a prediction horizon of \(N=20\). Fig.~\ref{fig:vehicle_analysis} illustrates the actual driving trajectories as well as the corresponding velocity and angular velocity profiles during real-world operation. Additionally, Fig.~\ref{fig. vehicle experiments} presents snapshots from the vehicle's navigation process. It is evident that when approaching an obstacle, TransMPC effectively adjusts its trajectory to steer around the obstacle, consistently maintaining a safe clearance. The resulting control commands remain smooth and within the actuator constraints. These real-world experiments demonstrate the effectiveness, practicality, and robustness of the proposed TransMPC policy, highlighting its suitability for deployment in dynamic environments on resource-constrained mobile robot platforms.

\section{Conclusion}
\label{sec: conclusion}

This paper introduces TransMPC, a novel explicit MPC method designed for real-time control of nonlinear systems, capable of adapting seamlessly to varying prediction horizons. The core innovation is formulating the MPC policy as an encoder-only Transformer network, utilizing bidirectional self-attention to directly map current states and reference trajectories of arbitrary lengths into complete action sequences within a single forward inference step. This approach inherently supports flexible horizon lengths and significantly reduces inference latency through parallel processing. TransMPC employs a direct policy optimization algorithm that alternates between sampling and training, optimizing the true finite-horizon MPC cost via automatic differentiation without relying on precomputed optimal trajectories. A replay buffer and random horizon sampling strategy guarantee robust generalization across diverse states and prediction horizons. Simulation studies on vehicle trajectory tracking tasks confirm that TransMPC achieves superior accuracy compared to baselines utilizing MLP and recurrent architectures, maintaining high performance across prediction horizons ranging from 1 to 20 steps. Real-world experiments involving trajectory tracking and obstacle avoidance further validate the practical effectiveness and robustness of TransMPC, highlighting its suitability for industrial applications. Future work will focus on extending TransMPC to handle non-differentiable cost functions and broader application scenarios in robotics and autonomous control systems.

% \section{}
% \label{Appendix}

% \vspace{\baselineskip}
% \textit{Proof.} See Appendix \ref{Appendix} in the supplementary material.
% \vspace{\baselineskip}

\ifCLASSOPTIONcaptionsoff
  \newpage
\fi

% References
\bibliographystyle{IEEEtran}
\bibliography{ref}
%IEEEabrv instead of IEEEfull

\vskip -2\baselineskip plus -1fil
\begin{IEEEbiography}[{\includegraphics[width=1in,height=1.25in,clip,keepaspectratio]{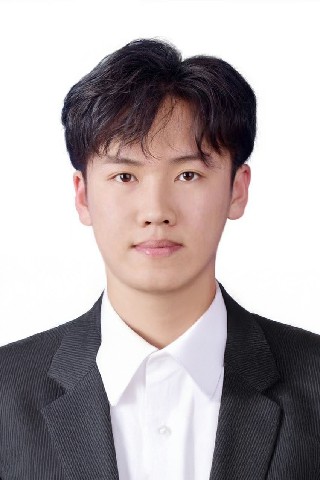}}]
{Sichao Wu} received the B.S. degree in vehicle engineering from the Yanshan University in 2023. He is currently pursuing an M.S. degree in vehicle engineering at the University of Science and Technology Beijing. 
His research interests include reinforcement learning, optimal control, and self-driving decision-making.
\end{IEEEbiography}

\vskip -2\baselineskip plus -1fil
\begin{IEEEbiography}[{\includegraphics[width=1in,height=1.25in,clip,keepaspectratio]{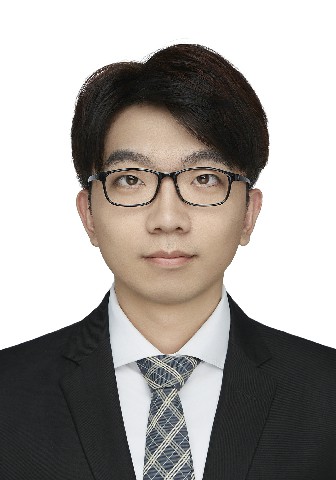}}]{Jiang Wu} received the B.S. degree in the School of Mechanical Engineering from Tianjin University of Technology, China, in 2022. He is currently pursuing an M.S. degree in Mechanical Engineering at the University of Science and Technology Beijing. 
His research interests include optimal control and kinematics analysis of robot manipulator.
\end{IEEEbiography}

\vskip -2\baselineskip plus -1fil
\begin{IEEEbiography}[{\includegraphics[width=1in,height=1.25in,clip,keepaspectratio]{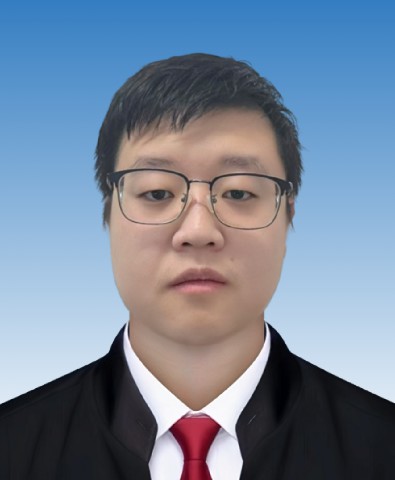}}]{Xingyu Cao} received the B.S. degree in mechanical engineering from Hefei University of Technology, Hefei, China. He is currently pursuing the Ph.D. degree at the University of Science and Technology Beijing, Beijing, China. His research interests include reinforcement learning and optimal control applications in articulated vehicles.
\end{IEEEbiography}

\vskip -2\baselineskip plus -1fil
\begin{IEEEbiography}[{\includegraphics[width=1in,height=1.25in,clip,keepaspectratio]{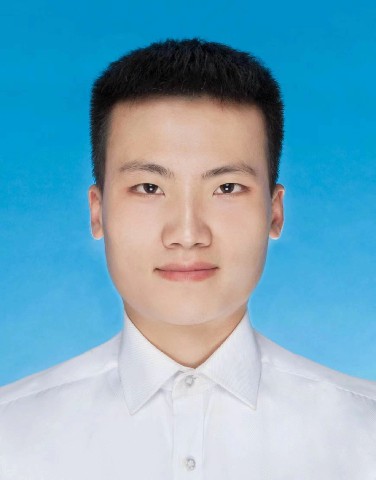}}]{Fawang Zhang} received the Master's degree in vehicle engineering from China Agricultural University, China, in 2019. He is currently a Ph.D. candidate in vehicle engineering at School of Mechanical Engineering, Beijing Institute of Technology. 
 He studied as a joint student at Tsinghua iDLab during 2019-2022. His active research interests include intelligent vehicle's planning and control, reinforcement learning, and optimal control.
\end{IEEEbiography}

\vskip -2\baselineskip plus -1fil
\begin{IEEEbiography}[{\includegraphics[width=1in,height=1.25in,clip,keepaspectratio]{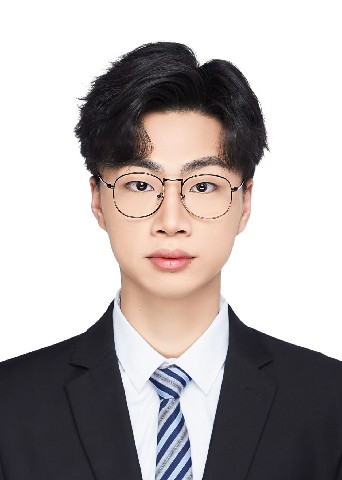}}]{Guangyuan Yu} received the B.S. degree in vehicle engineering from the University of Science and Technology Beijing in 2024.He is currently pursuing an M.S. degree in mechanical engineering at the University of Science andTechnology Beijing.His research interests include reinforcement learning,optimal control,and selfdriving decision-making.
\end{IEEEbiography}

\vskip -2\baselineskip plus -1fil
\begin{IEEEbiography}[{\includegraphics[width=1in,height=1.25in,clip,keepaspectratio]{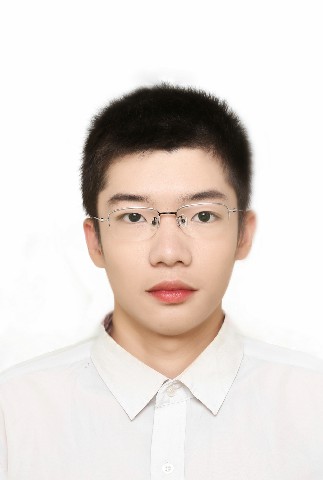}}]{Junjie Zhao} is currently pursuing an M.S. degree in mechanical engineering at the University of Science and Technology Beijing. 
His research interests include reinforcement learning, optimal control, and self-driving decision-making.
\end{IEEEbiography}

\vskip -2\baselineskip plus -1fil
\begin{IEEEbiography}[{\includegraphics[width=1in,height=1.25in,clip,keepaspectratio]{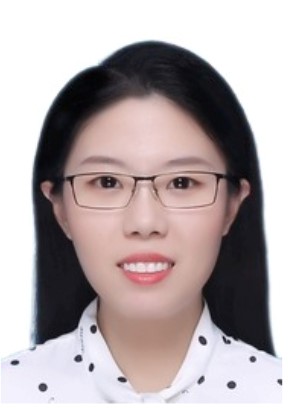}}]{Yue Qu} received her B.S. degree in Mechanical Engineering from Nanjing University of Science and Technology, China, in 2017 and her Ph.D. degree in Control Science and Engineering from Nanjing University of Science and Technology, Nanjing, China, in 2023. She is currently a lecturer at Nanjing University of Industry Technology. Her research interests include model predictive control and reinforcement learning with applications to unmanned aerial vehicles.
\end{IEEEbiography}

\vskip -2\baselineskip plus -1fil
\begin{IEEEbiography}[{\includegraphics[width=1in,height=1.25in,clip,keepaspectratio]{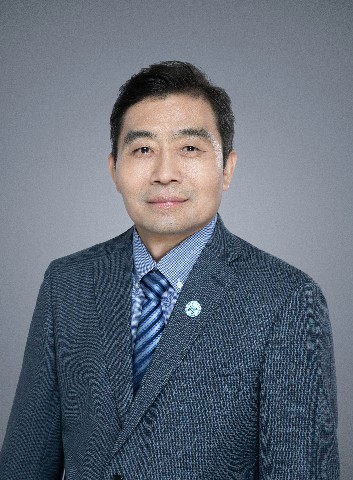}}]{Fei Ma} received the M.S. and Ph.D. degrees in mechanical engineering from the University of Science and Technology Beijing, Beijing, China, in 1993 and 2005. He is currently a Tenured Professor and Dean with the school of mechanical engineering, University of Science and Technology Beijing. His active research interests include vehicle dynamics, condition Monitoring, and fault diagnosis of machinery.
\end{IEEEbiography}

\vskip -2\baselineskip plus -1fil
\begin{IEEEbiography}[{\includegraphics[width=1in,height=1.25in,clip,keepaspectratio]{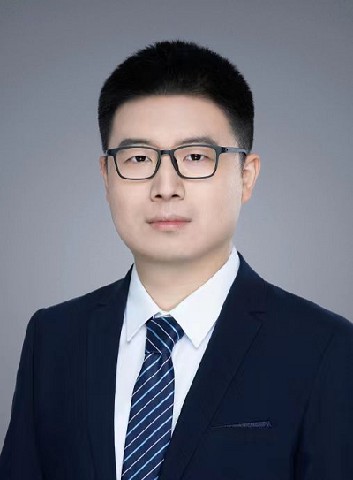}}]{Jingliang Duan} received his Ph.D. degree in the School of Vehicle and Mobility, Tsinghua University, China, in 2021.  He studied as a visiting student researcher in the Department of Mechanical Engineering, University of California, Berkeley, in 2019, and worked as a research fellow in the Department of Electrical and Computer Engineering, National University of Singapore, in from 2021 to 2022. He is currently an associate professor in the School of Mechanical Engineering, University of Science and Technology Beijing, China. His research interests include reinforcement learning, optimal control, and self-driving decision-making.
\end{IEEEbiography}
\end{document}